%% file: ms.tex
\documentclass[final,12pt]{clear2023} 

\newcommand{\newparagraph}[1]{\textbf{#1}\;\;}

\newcommand{\E}{\mathbb{E}}
\newcommand{\bX}{\mathbf{X}}
\newcommand{\bx}{\mathbf{x}}

\newcommand{\Var}{\textrm{Var}}

\newcommand{\bg}{\mathbf{g}}
\newcommand{\bG}{\mathbf{G}}

\newcommand{\bY}{\mathbf{Y}}

\newcommand{\bT}{\mathbf{T}}

\newcommand{\bZ}{\mathbf{Z}}

\newcommand{\btheta}{\boldsymbol{\theta}}

\newcommand{\bTheta}{\boldsymbol{\Theta}}

\newcommand{\bM}{\mathbf{M}}
\newcommand{\bm}{\mathbf{m}}

\newcommand{\bp}{\textbf{p}}
\newcommand{\bP}{\textbf{P}}

\newcommand{\itspace}{\hspace{0.3cm}}
\usepackage{fancyvrb}

\usepackage{tikz}
\usetikzlibrary{bayesnet}
\usetikzlibrary{arrows}
\usepackage{color}
\usepackage{comment}
\usepackage{graphicx}
\usepackage{caption}
\usepackage{sidecap}
\usetikzlibrary{backgrounds}

\title[Image-based Treatment Effect Heterogeneity]{Image-based Treatment Effect Heterogeneity}
\usepackage{times}

\clearauthor{%
 \Name{Connor T. Jerzak} \Email{connor.jerzak@austin.utexas.edu}\\
 \addr Department of Government 
  \\ The University of Texas at Austin
 \\ \texttt{ConnorJerzak.com}
 \AND
 \Name{Fredrik Johansson} \Email{fredrik.johansson@chalmers.se}\\
 \addr 
 Data Science and AI Division
 \\ Chalmers University of Technology
 \\  \texttt{fredjo.com}
 \AND
 \Name{Adel Daoud} \Email{adel.daoud@liu.se}\\
 \addr Institute for Analytical Sociology 
 \\ Linköping University
 \\ \texttt{AdelDaoud.se}, \texttt{global-lab.ai}
}

\begin{document}

\maketitle

\begin{abstract}%
Randomized controlled trials (RCTs) are considered the gold standard for estimating the Average Treatment Effect (ATE) of interventions. One important use of RCTs is to study the causes of global poverty---a subject explicitly cited in the 2019 \emph{Sveriges Riksbank Prize in Economic Sciences in Memory of Alfred Nobel} awarded to Duflo, Banerjee, and Kremer “for their experimental approach to alleviating global poverty.” Because the ATE is a population summary, researchers often want to better understand how the treatment effect varies across different populations by conditioning on tabular variables such as age and ethnicity that were measured during the RCT data collection. Although such variables carry substantive importance, they are often only observed only near the time of the experiment: exclusive use of such variables may fail to capture historical, geographical, or neighborhood-specific contributors to effect variation. In global poverty research, when the geographical location of the experiment units is approximately known, satellite imagery can provide a window into such historical and geographical factors important for understanding heterogeneity. However, there is no causal inference method that specifically enables applied researchers to analyze Conditional Average Treatment Effects (CATEs) from images. 
In this paper, we develop a deep probabilistic modeling framework that identifies clusters of images with similar treatment effect distributions, enabling researchers to analyze treatment effect variation by image.
Our interpretable image CATE model also emphasizes an image sensitivity factor that quantifies the importance of image segments in contributing to the mean effect cluster prediction.
We compare the proposed methods against alternatives in simulation; additionally, we show how the model works in an actual RCT, estimating the effects of an anti-poverty intervention in northern Uganda and obtaining a posterior predictive distribution over treatment effects for the rest of the country where no experimental data was collected. We make code for all modeling strategies available in an open-source software package and discuss their applicability in other domains (such as the biomedical sciences) where image data are also prevalent.
\end{abstract}
\vspace{0.1cm}

\clearpage \newpage 
\section{Introduction}\label{s:intro}
Field experiments in the social and health sciences help us understand the effects of interventions in the natural habitat where people live \citep{banerjee2011poor}. Their primary goal is often to identify the overall effect of a treatment $T_i$ on an outcome $Y_i$, marginalizing over units ($i\in\mathcal{I}$) in the sample population, and thereby, calculating the Average Treatment Effect (ATE). By collecting tabular characteristics, $\bX_i$, such as age and ethnicity, investigators can unpack results by sub-populations, estimating the Conditional Average Treatment Effects (CATEs) \citep{kunzel2019metalearners,balgiPersonalizedPublicPolicy2022}. However, features in $\bX_i$ are often measured only at baseline, just before the experiment is initiated. Thus, $\bX_i$ rarely contain information on an experimental unit's historical characteristics, including its past neighborhood-level features and geographical context, features which may be useful in identifying sub-populations that react differently to the same treatment \citep{kinoScopingReviewUse2021}. 

When experimental units $i$ are associated with a particular location, satellite images, $\bM_i$, can fill in this gap, providing important information about the otherwise missing historical, neighborhood, and geographical contexts \citep{burkeUsingSatelliteImagery2021,daoudUsingSatellitesArtificial2021a}. 

Indeed, in contrast to covariates measured during experiments, satellite imagery is collected passively from space for every geographic context on earth, and thus (except for clouds covering the line sight of a satellite) there is no systematic missingness in the data source. Moreover, these data have been collected for parts of the world since the CORONA intelligence satellites of the 1950s and for the entire world since the start of NASA and the US Geological Survey's joint Landsat mission in the 1970s. The Landsat data are publicly available with a revisiting time of on average 16 days. Therefore, by combining satellite imagery with experimental data, researchers can investigate not only the historical and geographical roots of effect variation, but they can also predict how an intervention will likely impact places outside the scope of the original study where no researcher-collected covariate data are available. 
That is, predictive distributions over treatment effects can be estimated not only for the experimental context but also investigated in places not originally conceived of during the experiment---where no tabular background covariates were measured. In this way, earth observation data have the potential to increase the applicability of ideas in casual transportability \citep{pearl2022external}. 

However, despite the potential offered by images for causal inference, as evident by the growing literature \citep{castroCausalityMattersMedical2020,chalupkaMultiLevelCauseEffectSystems2016,chalupkaUnsupervisedDiscoveryNino2016,chalupkaVisualCausalFeature2015,scholkopfCausalRepresentationLearning2021,yiCLEVRERCoLlisionEvents2020,dingDynamicVisualReasoning2021,paciorek2010importance,kaddourCausalEffectInference2021,louizosCausalEffectInference2017a,pawlowskiDeepStructuralCausal2020,lopez-pazDiscoveringCausalSignals2017a}, it remains unclear how researchers should use these $\bM_i$'s for CATE analysis. One key challenge is that $\bM_i$ is high-dimensional and rarely annotated. A CATE analysis using $\bM_i$ but relying on tabular methods, such as linear models or the generalized random forest \cite{athey2019generalized}, would likely lead to poor interpretability or struggle to find heterogeneity within the high-dimensional image tensors. 

For these raesons, there is a major research need to form a causal inference framework for CATE analysis in images. Equipped with such a method, applied researchers would not only be enabled to start incorporating satellite images in future RCTs, but also re-analyze data from past experiments that have already been performed, with the goal of deepening understanding using satellite image data, potentially uncovering missed yet significant CATE--informing policymaking.

In this article, we develop machine learning models which characterize image-based effect heterogeneity in the RCT setting. We introduce an interpretable CATE model that employs Bayesian convolutional neural network arms (CNNs) with categorical gates that allow us to directly model mixtures of image clusters with similar effects. Our models estimate treatment effects for all units, conditional on treatment status, the images $\bM_i$, and, if desired, accounting for available $\bX_i$ by incorporating them into the cluster prediction or via orthogonalization. By residualizing, our model will identify what additional CATE that stems from $\bM_i$, separately from $\bX_i$, for enhanced heterogeneity analysis. The models construct image-type clusters that group units probabilistically based on their CATE similarity. 

In the following sections, we develop our methods, show some of their properties analytically, and explore others in simulated experiments. We demonstrate the usefulness of our methods by replicating the results of an RCT study in Uganda---the Youth Opportunities Program (YOP) \citep{blattman2014generating}. This study, conducted in 2008, was designed to help the poor break unemployment cycles by financially assisting their artisans and business activity. The government solicited young adults to participate in YOP, asking them to form teams and compose a business plan. After screening teams, the government randomly assigned some to receive one-time grants worth about \$7,500, often more than members' joint annual income. Most of the applicants were young, rural farmers having low educational attainment ($\sim$ eighth grade), earning less than \$1 per day, and working less than twelve hours per week. In many RCTs like YOP, the researchers collected a set of baseline covariates, but none of them explicitly capture historical neighborhood or geographical characteristics. Our replication uses satellite imagery collected independently of the original experiment and demonstrates the usefulness, and limitations, of using $\bM_i$ for CATE, as a complement to the tabular version. 

While our contribution focuses on the use of satellite images in global poverty research, our methods are designed such that they generalize to other RCT settings where complementary image data are available. In the last few decades, there has been a rapid increase in the availability of imaging technologies. Most notably, these technologies are readily available in biomedical fields in the form of X-ray, positron emission, MRI, and ultrasound modalities. These images data streams are likely useful not only for estimating ATE \citep{castroCausalityMattersMedical2020, lopez-pazDiscoveringCausalSignals2017a, chalupkaMultiLevelCauseEffectSystems2016}, but also for evaluating effect heterogeneity. More research will be needed to determine the usefulness of our modeling approach for such domains.

\section{Background and Related Work}\label{s:lit}
\newparagraph{CATEs with Tabular Data}
Let $Y_i(t)$ denote the potential outcome~\citep{rubin2005causal} of an intervention $t \in \{0,1\}$ for a unit of study $i$. For example, $Y_i(1)$ may represent the poverty level in household $i$ following an aid intervention, and $Y_i(0)$ is the level without intervention. 

We may define the unit-level treatment effect as $\tau_i = Y_i(1) - Y_i(0).$ When $\tau_i$ is greater than 0, the unit's outcome is greater under treatment than otherwise. The quantity $\tau_i$ cannot be exactly identified without strong assumptions \citep{pearl2009causality}. Because a unit can only receive a single treatment at a given time, only one of the potential outcomes, $Y_i(1)$ or $Y_i(0)$, is observed, and thus, the counterfactual remains unobserved. Assuming consistency---that is, units comply with their treatment assignment---the observed outcome can be written as, $Y_i = Y_i(T_i) = Y_i(1) T_i + Y_i(0) (1 - T_i)$, where $T_i$ denotes the (random) treatment status of $i$ \citep{hernanbook}. The ATE captures the population effect by averaging over all unit-level effects:
\[
\textit{Average Treatment Effect (ATE):\itspace} \E[\tau_i] = \E[Y_i(1) - Y_i(0)].
\]
The ATE is useful as it marginalizes over the heterogeneity present in a population to form an overall assessment of an experiment. With treatment randomization, ATE can be estimated non-parametrically by the difference between treatment and control outcomes~\citep{rubin2005causal}. Despite the importance of aggregate quantities such as the ATE, it is useful to disaggregate average effects using based on contextual information. Such a disaggregation is critical for not only scientific understanding but also for policy learning (e.g., by personalizing treatments \citep{greenland2020causal,balgiPersonalizedPublicPolicy2022}. This disaggregation can be a function of any type of general pre-treatment data variable, $\bG_i$: 
\begin{align*} 
\textit{Conditional Average Treatment Effect (CATE):\itspace} \tau(\bg) = \E[Y_i(1) - Y_i(0)\mid\bG_i=\bg],
\end{align*} 
The literature has primarily focused on conditioning sets that contain tabular data, $\bX_i$: 
\begin{align*} 
\textit{Tabular Conditional Average Treatment Effect:\itspace} \tau(\bx) = \E[Y_i(1) - Y_i(0)\mid\bX_i=\bx],
\end{align*} 
where $\bx \in \mathbb{R}^D$ denotes the vector of $D$ pre-treatment covariates \citep{athey2016recursive,athey2019generalized,ding2016randomization,imai2013estimating,shalit2017estimating,zhao2017selective,luedtke2016super,nie2021quasi}. 
For example, the generalized random forest is one such machine-learning method  \citep{athey2019generalized}, and it has proven useful in a variety of applied settings \citep{shiba2021heterogeneity,daoud2019estimating}. However, these methods are tailored for annotated tabular data and images are high-dimensional, often non-annotated. These non-annotated image features consist of image bands and pixels that may jointly induce effect heterogeneity. Thus, more research is required to improve the ability of investigators to understand CATE in the context of unstructured high-dimensional image data. 

\newparagraph{Causal Inference with Image Data}
While most causal-inference studies use tabular data, there is an increasing realization that image data provides a creative yet useful way to conduct causal inference \citep{castroCausalityMattersMedical2020,ramachandra2019causal,daoudStatisticalModelingThree2020}. To this end, there is a growing methodological literature investigating how images should be integrated to identify and estimate ATEs in the observational setting \citep{kallus2020deepmatch,kaddour2021causal,pawlowskiDeepStructuralCausal2020,jerzak2023integrating}. 
Yet these approaches tend to mainly treat images as proxies, for inclusion in the adjustment set, thereby securing causal identification; these approaches are not tailored for CATE analysis in images. Hence, little is known about how to use images for CATE. 

Like tabular information in $\bX_i$, images as a whole or through its segments may be associated with treatment effect heterogeneity. In the observational setting, images $\bM_i$ could be part of both the conditioning and effect modification sets. In the RCT setting, $\bM_i$ is not a confounder, since the treatment was randomized, but it  may  provide significant information for CATE (as discussed in \S\ref{s:intro}). In both settings, one target estimand is the Image CATE,
\[
\textit{Image CATE:\itspace} \tau(\bm) = \E[Y_i(1) - Y_i(0) \mid \bM_i=\bm], 
\]
where $\bm\in \mathbb{R}^{W\times H \times D}$ is an image, obtained before treatment assignment, of width $W$, height $H$, and with $D$ data channels. These data channels often contain reflectance information from various electromagnetic bands. 
In some applications involving earth observation data, the $i$ subscript may correspond to a spatially defined neighborhood (or to a person living in such a place). In applications involving medical imaging, the $i$ subscript may correspond to patients or tissue regions. In both, the Image CATE analysis will have a connection with Multiple Instance Learning (MIL) in the sense that a single effect response (e.g., high or low) may be associated with multiple image segments (for a review of MIL, see \citet{foulds2010review}).

Although image and tabular CATEs are both special cases of the general CATE, there are conceptual differences between them. First, images are unstructured, high-dimensional objects, and satellite images of each neighborhood are unique in RCT applications. This makes it difficult, if not impossible, to perform non-parametric inference. Second, since images are unstructured, it is unclear how to interpret the act of conditioning on an image. We need a conceptual language and modeling strategy for characterizing proximity between images in the space of conditional effects. Therefore, the remainder of our article will contribute to establishing this conceptual language for RCTs, leaving image CATE in the observational case for future study. 

\section{Modeling Causal Effect Heterogeneity in Images}
We first introduce a baseline method of interpretable image CATE analysis which we will later contrast against the probabilistic Image-Type Effect Cluster Model, which will form the focus of our later application. 

\subsection{Comparative Baseline: Prediction Cluster CATE}\label{s:BaselineModel}
In experimental settings, CATEs may be estimated readily by function approximation. Perhaps the simplest approach is to use a parameterized function, $f_{\hat{Y}_t}(\bm)$, to predict potential outcomes, $Y_i(t)$, of each intervention---a so-called \emph{T-learner}~\citep{kunzel2019metalearners}. The CATE may then be estimated as $\hat{\tau}(\bm) = f_{\hat{Y}_1}(\bm)-f_{\hat{Y}_0}(\bm)$. \citet{shalit2017estimating} found that learning a shared representation used to predict both treatment outcomes improved prediction quality and named this approach, {\it TARNet}. 

Given a CATE model formed by such function approximation, we can aim to increase the interpretability of the model output by partitioning units by their predicted causal effect, creating a clustering of inputs \emph{post hoc}. This post-hoc clustering will serve as a comparative baseline for the subsequent probabilitistic models. Let $f_{\textrm{Cluster}}(\hat{\tau}(\bm)) \in \{1, 2, ..., K\}$ denote a cluster labeling function determined by the output of $\hat{\tau}$, which partitions the space of effect sizes and, as a result, the space of images. $C$ may be constructed by quantile binning of $\hat{\tau}$, or, as in our experiments, by $k$-means clustering. The CATE with respect to the post-hoc clustering labels is
\begin{align*} 
& \textit{Prediction Cluster CATE:\itspace} \tau(c) = \E[Y_i(1) - Y_i(0) \mid f_{\textrm{Cluster}}(\hat{\tau}(\bM_i)) = c],
\end{align*} 
In our experiments, we use the post-hoc clustering of the $\hat{\tau}(\bM_i)$'s from TARNet as a point of comparison, as the approach is a standard one for high-dimensional causal estimation. 

A drawback of a post-hoc approach is that it compounds approximations to arrive at a discrete representation of treatment effect heterogeneity. If two similar images yield very different predictions due to misestimation in either the model for $Y_i(0)$ or for $Y_i(1)$, they are likely to be placed in different clusters. We would prefer to cluster images in a way that smoothly best approximates the Image CATE in a single model. Moreover, it is difficult to quantify how the image affects the  post-hoc clustering because it may be computationally prohibitive to propagate gradients through both the outcome and subsequent clustering model (a topic explored in \S\ref{ss:sensitivity}). 

To address these limitations, we therefore develop a probabilistic image CATE model that directly targets the prediction of the heterogeneity itself in a low-dimensional summary, with the goal of increasing the interpretability of the image heterogeneity dynamics. 

\subsection{Interpretable Models for Effect Clustering Based on Image Type}\label{s:ProbModel}
We now introduce a series of modeling strategies for directly targeting CATE clusters in images; to the best of our knowledge, this is the first work to use images to estimate CATE with a particular focus on interpretability. 
We will aim to fulfill the following criteria: (1) \emph{model potential outcomes and treatment effects flexibly (e.g., allowing for non-linearities)} (2) \emph{identify interpretable image clusters with similar in-cluster effect sizes, and different cross-cluster effects}, and (3) \emph{allow for the modeling of uncertainty regarding the image clusters and treatment effects}. 

Our target quantity of interest will be
\begin{align*} 
& \textit{Image-Type CATE:\itspace} \tau(z) = \E[Y_i(1) - Y_i(0) \mid Z_i = z],
\end{align*} 
where $Z_i\in \{1,2,...,K\}$ denotes the effect cluster of image $\bM_i$, where there are $K$ total clusters. We will search for assignments $Z_i$ of clusters to images that best explain treatment effect variation.\footnote{A related quantity is targeted in the mixture-of-experts approach for CATEs in the conjoint setting using linear models with interactions \citep{goplerud2022estimating}. }

Because the treatment effects are decomposed by image type, there are only $K$ quantities needed to effectively summarize the heterogeneity attributable to images. Since human working memory can track around 5 distinct chunks at a time \citep{cowan2010magical}, this low-dimensional probabilistic summary of the complex heterogeneity process stemming from images can in principle be communicated to human stakeholders---thereby facilitating future treatment-targeting decisions. With this substantive motivation in mind, we now discuss how we meet the modeling  criteria for capturing treatment effect heterogeneity with images. 

\begin{SCfigure}
  \caption{A stylized schematic depiction of the probabilistic treatment heterogeneity model for images. Gray circles denote observed random variables;  white circles denote latent variables. Mixed gray/white nodes denote partially observed nodes (i.e., nodes observed for some, but not all, units). Square nodes denote deterministic transformations. $Z_i$ denotes the image type generating a distribution over treatment effects. Arrows denote statistical (not causal) dependencies. \label{fig:model}}
  
\includegraphics[width=0.55\linewidth]{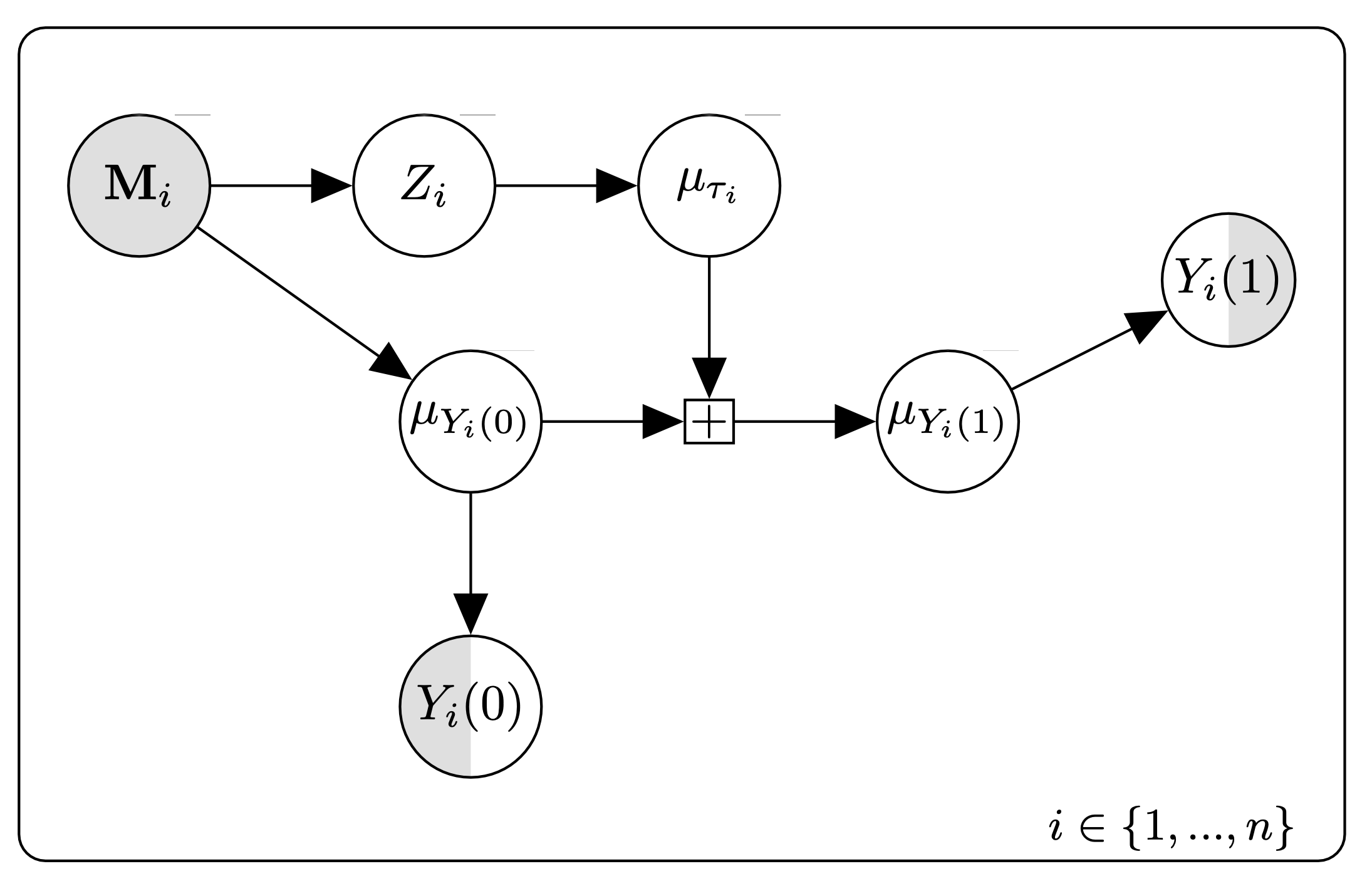}

\end{SCfigure}

\newparagraph{Probabilistic Estimation of the Baseline Outcome}
We satisfy the first criterion by allowing the baseline potential outcome, $Y_i(0)$, to be estimated flexibly as a function of the image. In particular, we let the mean of the baseline potential outcome, conditional on the image, be parameterized by a Bayesian convolutional neural network,
\begin{equation}\label{eq:BCNN}
\E[Y_i(0) \mid \bM_i = \bm]  \ = \ \{ \mu_{Y_i(0)} \mid \bM_i = \bm \} \ \sim\ \textrm{Bayesian CNN}(\bm) ~,
\end{equation}
where convolutional and dense parameters are not deterministic but instead defined according to a distribution. This model for the baseline mean is listed as (2) in Schema 1 and depicted as the arrow between $\bM_i$ and $\mu_{Y_i(0)}$ in the probabilistic model depiction in Figure \ref{fig:model}. 

\newparagraph{ Effect Mixture Based on Image Type}
Having defined the baseline, we next turn our attention to the modeling component that targets the Image-Type CATE estimand. In this component, we first compute image type probabilities $\bP_i$ using another Bayesian CNN. Given $P_{iz}=p_{z}$, the image takes on cluster type $z$ with probability $p_z$. Intuitively, while the first CNN looks for image patterns indicative of $\mu_{Y_i(0)}$, the second looks for patterns associated with the type of treatment response. We develop two model variants for the type response characterization---one more interpretable and the other more flexible.

\newparagraph{{\it Variant 1.} Image-Type Effect Cluster Model}
Here, conditional on the \emph{image type} $Z=z$, mean treatment effects are drawn from a Normal with a mean $\mu_{\tau, z}$ and variance $\sigma^2_{\tau, z}$  indexed to that image type. We do not assume that there is a single treatment effect per image type, but instead that there is a specific \emph{distribution} over treatment effects by image type. The cluster effect means and variances offer a complete summarization of this distribution. This model emphasizes interpretability: given the image type, treatment effects are characterized by a single, unique distribution. 

\newparagraph{ {\it Variant 2.} Image-Type Differential Effect Model}\label{ss:DifferentialImageType}
The first effect mixture model could be useful to investigators because conditional effects can be succinctly summarized with a small number of parameters, but in some circumstances, investigators may want a more flexible model for the heterogeneity structure. In that case, the distribution of treatment effects given the image type $z$ is parameterized by a Bayesian CNN arm indexed to $z$: 
\[
\{\mu_{\tau_i} \mid \bM_i=\bm, \ Z_i = z \} \sim \textrm{Bayesian CNN}_z(\bm)
\]
Here, the image type acts as a stochastic gate that determines which image patterns will be used in predicting the mean treatment effect value given the image, $\bM_i$. 

Overall, the probabilistic generative modeling framework for image-based CATE is summarized visually in Figure \ref{fig:model} and as follows:
{
\vspace{-0.1cm}
\begin{center} {Schema 1:} 
The Image Effect Cluster Model. See Figure \ref{fig:model} for visualization.
\vspace{-0.25cm}  \end{center}
\hrulefill
\footnotesize
  \begin{eqnarray*}
& \text{(1a) }\textit{\underline{Generating the Image Type}}
\\ &  \{ \bP_i \mid\bM_i=\bm \}  \sim \textrm{Bayesian CNN}(\bm)
\\ &  \{  Z_i \mid \bP_i=\bp \} \sim \textrm{Categorical}(\bp)
\\ & \downarrow
\\  &  \text{(1b) }\textit{\underline{Generating the Treatment Effect Distributions}}
\\ & \textrm{i. Image-Type Effect Cluster Model: } \{ \mu_{\tau_i} \mid Z_i = z \} \sim
\mathcal{N}( \mu_{\tau, z}, \ \sigma_{\tau, z}^2)
\\ & \textrm{ii. Image-Type Differential Effect Model: } \{ \mu_{\tau_i} \mid \bM_i=\bm, \ Z_i = z \} \sim \textrm{Bayesian CNN}_{z}(\bm)
\\ & \qquad\qquad\qquad\qquad\qquad\downarrow
\\ & \begin{matrix}
  \text{(2) }\textit{\underline{Generating the Baseline Outcome, $Y_i(0)$}} & \text{(3) } \textit{\underline{Generating the Outcome under Treatment, $Y_i(1)$}}
\\ \{ \mu_{Y_i(0)} \mid \bM_i = \bm \} \sim \textrm{Bayesian CNN}(\bm)   & \hspace{-.2cm}\rightarrow \{Y_i(1) \mid \mu_{Y_i(0)}, \mu_{\tau_i}, \ \sigma_{1,z}^2\} \sim \mathcal{N}(\mu_{Y_i(0)}  + \mu_{\tau_i}, \ \sigma_{1,z}^2)
\\ \{ Y_i(0) \mid \mu_i(0), \ \sigma_{0,z}^2  \} \sim \mathcal{N}(\mu_{Y_i(0)}, \ \sigma_{0,z}^2) & 
\end{matrix} 
\end{eqnarray*}
\hrulefill\hrulefill
\vspace{0.2cm}
}

There are several advantages to these modeling strategies. There is improved interpretability from summarizing image-derived effect heterogeneity in $K$ discrete clusters. Moreover, under the Image-Type Effect Cluster Model, we can efficiently summarize the cluster effects (see \S\ref{ss:DistributionTau}): 
\begin{align*}
\tau(z) = \E[Y_i(1) - Y_i(0)\mid Z_i = z] &= \mu_{\tau,z}, \ \;\;\;\; \Var(Y_i(1) - Y_i(0)\mid Z_i = z) \ = \ \sigma_{0,z}^2+ \sigma_{1,z}^2 +\sigma_{\tau,z}^2.
\end{align*}
Next, as the strategies are probabilistic, so we can explore uncertainty not only in the image cluster effects by also in the cluster assignment probabilities.\footnote{The approach outlined here can be readily adapted to outcomes having non-Normally distributed outcomes by selecting different observed data likelihoods.} In addition, the cluster decomposition may facilitate scientific inquiry: an image type serves as a generalization tool for reasoning across images, facilitating theorizing about the causal mechanisms at play. Finally, we can readily compute the gradients of the expected cluster probabilities with respect to the image in order to identify \emph{how} the image affects the typology, a matter explored in \S\ref{ss:sensitivity}.

For both probabilistic model variants, estimation is performed via variational Bayesian methods\footnote{We note in passing that an additional benefit of the approach proposed here, as opposed to post-hoc clustering, is that uncertainty of the variational clustering model can be further quantified under model misspecification using $M$-estimation theory \citep{westling2019beyond}.}  where we learn the joint distribution of the model parameters $\bTheta$ and the image clustering, $\bZ$, given the observed dataset, $\mathbf{D}=\{Y_i(T_i), \ \bM_i, \ T_i \}_{i=1}^n$:
\begin{equation}\label{eq:posterior}
\textit{Target Posterior:\itspace}
p(\bZ, \ \bTheta \mid \mathbf{D}),
\end{equation}
where $\bZ=\{Z_i\}_{i=1}^n.$
For details of how we model the uncertainties, see \ref{ss:ModelDetails}. To estimate the posterior in \eqref{eq:posterior}, we maximize the Evidence Lower Bound (ELBO)~\citep{ranganath2014black},
\begin{align*}
\underset{q(\mathbf{Z},\  \bTheta)}{\textrm{maximize}} \itspace &\E_{q(\mathbf{Z},\ \bTheta)}\left[  \log\big(p(\bY(\bT) \mid \bZ, \ \bTheta, \  \bM, \ \bT)\big) \right]
\ -  \
D_{\textrm{KL}}\left(
q(\bZ, \ \bTheta)
\;||\; 
p(\bZ, \  \bTheta)
\right).
\end{align*}
We solve the problem approximately using stochastic gradient descent with gradients passing through discrete sampling nodes using re-parameterization techniques \citep{parmas2021unified}. The choice of priors affects finite sample performance; when possible, we specify priors using  observable marginal information (e.g., prior means for cluster effects are set to $\widehat{\E}[Y_i(1) - Y_i(0)]$).


\subsection{Determination of Salience Regions in the Posterior Mean Probabilities}\label{ss:sensitivity}
One benefit of the Bayesian image-type heterogeneity model is that we can examine the model in order to assess {\it how} image information translates into the predicted effect cluster. In particular, we can take the derivative of the posterior mean cluster probabilities with respect to pixel $(w,h)$: 
\begin{align*} 
s_{whk}^{\textrm{Direction}} =  \sum_{c=1}^C \frac{\partial \ \E\left[\Pr(Z_i = k \mid \bM_i=\bm)\right] }{ 
\partial \ m_{whc}},
\end{align*}
where $c\in\{1,...,C\}$ denotes the channel (band) dimension. The quantity, $s_{whk}$ is a scalar summary of how changing pixel $(w,h)$ would induce changes in the predicted effect cluster probability. Because the modeling strategy is probabilistic, the salience must average over the randomness in the predicted cluster probabilities, hence the expectation on the inside of the derivative. This expectation is approximated via Monte Carlo. Positive values of this quantity would indicate that increasing the pixel intensity at $(w,h)$ would increase the probability of cluster $k$; with the same logic, negative values indicate that increasing pixel intensity at $(w,h)$ would decrease the probability of cluster $k$. 

Because directional salience may not be interpretable when increasing the pixel intensity is not itself interpretable, we can also consider an approach based on magnitudes, bypassing the potentially difficult interpretation of pixel intensities in different bands: 
\begin{align*} 
s_{whk}^{\textrm{Magnitude}} =  \sqrt{\sum_{c=1}^C \left(\frac{\partial \ \E\left[\Pr(Z_i= k \mid \bM_i=\bm)\right] }{ 
\partial \ m_{whc}}\right)^2}.
\end{align*}
Large values of $s_{whk}^{\textrm{Magnitude}}$ reveal locations in the image that, if changed, would lead to large changes in cluster $k$ probabilities. This measure is agnostic about whether those changes would be associated with increases or decreases in those expected probabilities. This salience information is difficult to compute for post-hoc clustering methods, as the clustering of the $\hat{\tau}_i$'s needs to solve a second optimization problem (as in $k$-means) through which gradients with respect to the initial outcome model(s) may not be efficiently traced. Moreover, in contrast to the post-hoc approach of \S\ref{s:BaselineModel}, the salience measure here incorporates uncertainty in the cluster prediction itself (since the salience  averages over randomness in the cluster probabilities).


\subsection{Policy Action Using the Image-based Heterogeneity Model}
A major motivation for considering satellite-image-based CATEs is that we can readily generate predictive distributions over treatment effects for contexts outside the experimental setting and where no tabular covariates were measured by researchers---a possibility that may meaningfully expand the reach of causal analyses. In this context, for a new out-of-sample point not from the original dataset, $i^{\textrm{Out}} \in \mathcal{I}(\textrm{Out})$,  we form a predictive distribution over image treatment effects using 
\begin{align*}
p(\tau_{i^{\textrm{Out}}}
\mid \bM_{i^{\textrm{Out}}}, \ \mathbf{D}) &= \sum_{z=1}^K  \int p(\tau_{i} \mid  Z_{i^{\textrm{Out}}}=z ; \ \bTheta=\btheta)  
\cdot p(Z_{i^{\textrm{Out}}}=z \mid \bM_{i^{\textrm{Out}}}; \ \bTheta=\btheta) \cdot p(\bTheta=\btheta \mid \mathbf{D})\; \textrm{d} \btheta 
\end{align*} 
We can use the predictive distribution over treatment effects to improve treatment targeting for out-of-sample individuals. With a fixed treatment budget of size $n_1^{O}$ for the new dataset of size $n^{O}=|\mathcal{I}(\textrm{Out})|$, this policy can be written as $\boldsymbol{\Pi}(\{\bM_{i^{\textrm{Out}}}\}_{\mathcal{I}(\textrm{Out})}) \to\{0,1\}^{n^O}$.
There are many approaches to this problem, and we refer readers to the relevant literature \citep{hitsch2018heterogeneous}.\footnote{We also note that, for the predicted treatment effects to be reliable, there should be minimal distribution shift between in- and out-of-sample points. In practice, this means that the experimental areas should ideally be selected randomly from within the geographic unit of interest, so that extrapolations into data-sparse regions are minimal.}

\subsection{Multi-modal Learning with Image and Tabular Heterogeneity} 
Tabular information can be readily incorporated into this modeling pipeline. For example, tabular covariates can be appended to the input of the dense part of the cluster type and baseline outcome models. The resulting treatment effect heterogeneity clusters are therefore conditional on both individual-level and also image-context-level information. This image and tabular data integration can be useful when investigators are focused on understanding the holistic heterogeneity dynamic in an experimental context, integrating both individual- and neighborhood-level information. Such a combined approach, an example of multi-modal learning \citep{ullah2022review}, can also be potentially useful in the medical domain, where image and high-dimensional medical records text can form the basis for improved patient response modeling. 

\subsection{Distinguishing Image from Tabular Heterogeneity} 
When tabular covariates were not measured or when we aim to generate predictions for observations having no measured tabular covariates, it may be useful to perform image-type effect clustering directly using images alone. When other tabular covariates are measured for the experimental sample, researchers may seek to understand the heterogeneity stemming from image information that is unique when compared with tabular covariates. 

For example, information about economic class is embedded in earth observation images, since the urban poor are often concentrated in dense city centers while the affluent often live in green-space-rich areas outside cities \citep{venter2020green}. We may wonder about the remaining heterogeneity after accounting for tabular variables such as income. In this context, we can perform the image clustering on the orthogonalized outcomes: $Y_i(t)^{\perp} = Y_i(t) - \widehat{\E}[Y_i(t) | \bX_i]$. The resulting clusters can then be interpreted as image effect types after accounting for the additive heterogeneity from measured tabular data. This image-specific decomposition can be helpful when geographic or neighborhood information is the focus of study.

\section{Treatment Effect Cluster Recovery in Simulation}\label{ss:sim}
We now explore the dynamics of the proposed methods in simulation, where true treatment effects are known. We generate image-based treatment effect heterogeneity using, 
\begin{align}\label{eq:sim}
H_i = \textrm{GN}(\textrm{max}( f_l(\bM_i) ) ),
\qquad H_i^{+} = \underbrace{|\min\{H_i\}_{i=1}^n|}_{\color{black}\textrm{Ensures $\tau_i>0$}} + \underbrace{\textrm{sign}(H_i)\cdot |H_i|^{1/\gamma}}_{{\color{black}\textrm{Generates bimodality as $\gamma \to \infty$}}}
\end{align}
where $f_l(\cdot)$ denotes the application of a $l\times l$ filter to the image, max($\cdot$) denotes the global maximum operation across the image, and $\textrm{GN}(\cdot)$ denotes a global normalization function scaling the $H_i$ values to have mean 0 and variance 1 across the image pool. The specific transformation generating $H_i^+$ is selected to ensure all the treatment effects are in the same direction (i.e., all positive) and to generate heterogeneity in the effect distribution, with greater bimodality in the treatment response as $\gamma\to \infty$. We let $\gamma = 2$. We define the treatment and outcome: 
\begin{align*} 
\begin{matrix}
T_i \sim \textrm{Binomial}(0.5),
& Y_{i} = T_i H_i^+ + \epsilon_{i},
\end{matrix}
\end{align*} 
with $\epsilon_i \sim \mathcal{N}(0, \ \nu\cdot \textrm{Var}(H_i^+))$. The value of $\nu$ controls the extent to which the image is predictive of the outcomes, where smaller values indicate a stronger image heterogeneity signal. To explore the effect of the signal-to-noise ratio, we vary $\nu \in\{0.01,0.1,1\}$.

The filter used in the convolution function of Equation \ref{eq:sim} is visualized in Figure \ref{fig:SimIll}, along with high and low responders from the set of images used in the simulation that we take from Landsat mosaics of sub-Saharan Africa. We have some degree of model misspecification here, as the way the data are generated is distinct from the estimation models; given this, we will examine the degree to which the various models will recover key properties of the image-based causal system.

\newparagraph{Cluster Recovery Measure} We compare the estimated effect clustering with an oracle baseline from the true, in practice unknown, $\tau_i$'s. That is, we first compute the oracle $k$-means clustering:
\begin{align*} 
\tau(z)_{\textrm{Oracle}} = \textrm{$z^{\textrm{th}}$ center from the oracle $k$-means applied to the true (in practice unknown) $\tau_i$'s}
\end{align*} 
The clustering quality measure then compares the oracle with estimated cluster means: 
\begin{align*} 
\textrm{Cluster Recovery $R^2$} \ \ &= \ \ 1 - \frac{ \sum_{z=1}^K \min_{z'} (\widehat{\tau}(z) - \tau(z')_{\textrm{Oracle}})^2 }{
\sum_{z''=1}^K (\tau(z'')_{\textrm{Oracle}}  - \overline{\tau}_{\textrm{Oracle}})^2},
\end{align*} 
where $\overline{\tau}_{\textrm{Oracle}}$ denotes the mean across the oracle cluster centers. This measure is equivalent to the $R^2$ in predicting the oracle cluster means from the estimated ones, where the ordering of the clusters has been arranged so that each oracle center is compared to its nearest estimated counterpart. 

\begin{SCfigure}
\caption{\emph{Left:} Capturing the treatment effect heterogeneity with our Image-Type Effect Model. \emph{Right:} Comparing models as we vary the signal-to-noise ratio.}\label{fig:SimResults}
\includegraphics[width=0.29\linewidth]{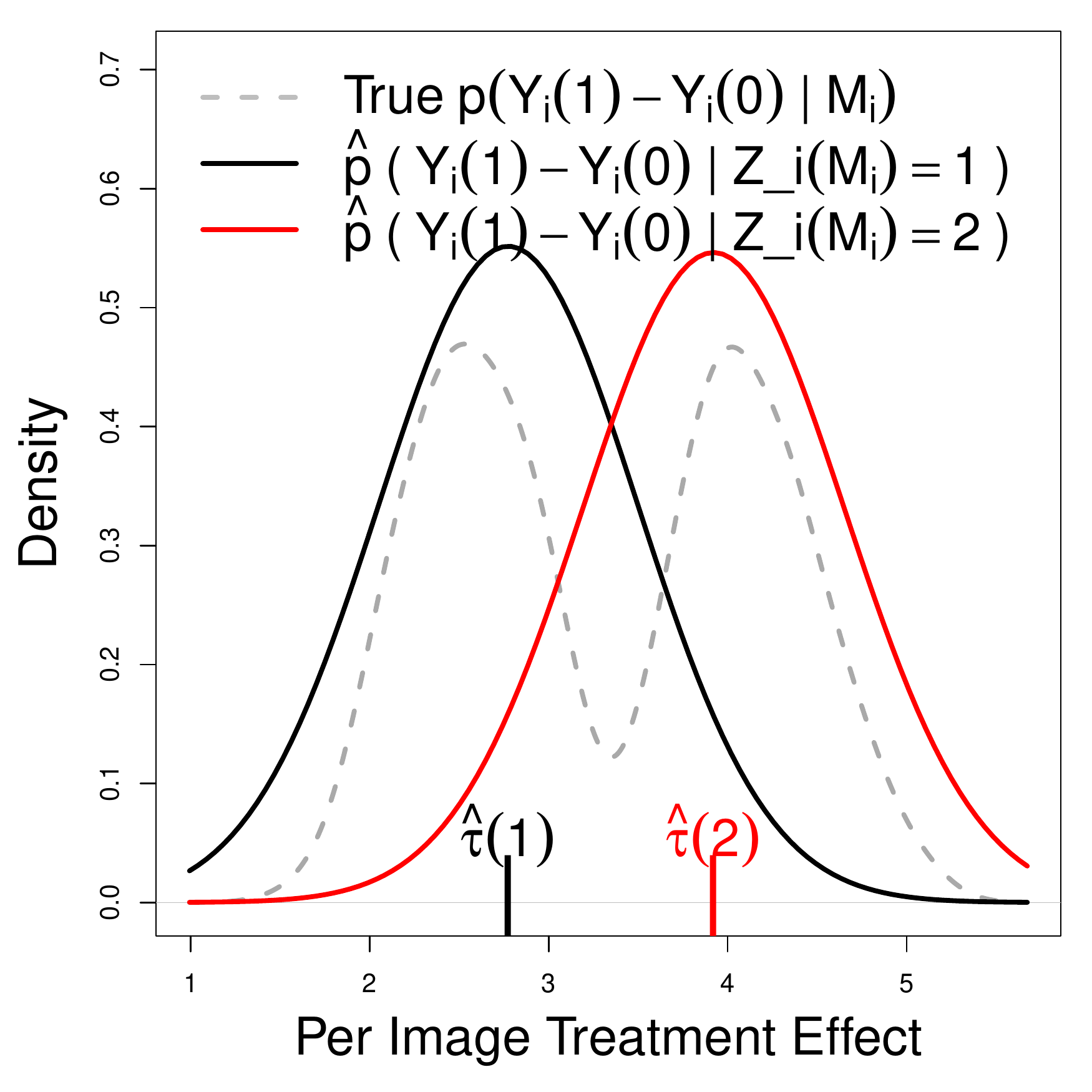}
\includegraphics[width=0.29\linewidth]{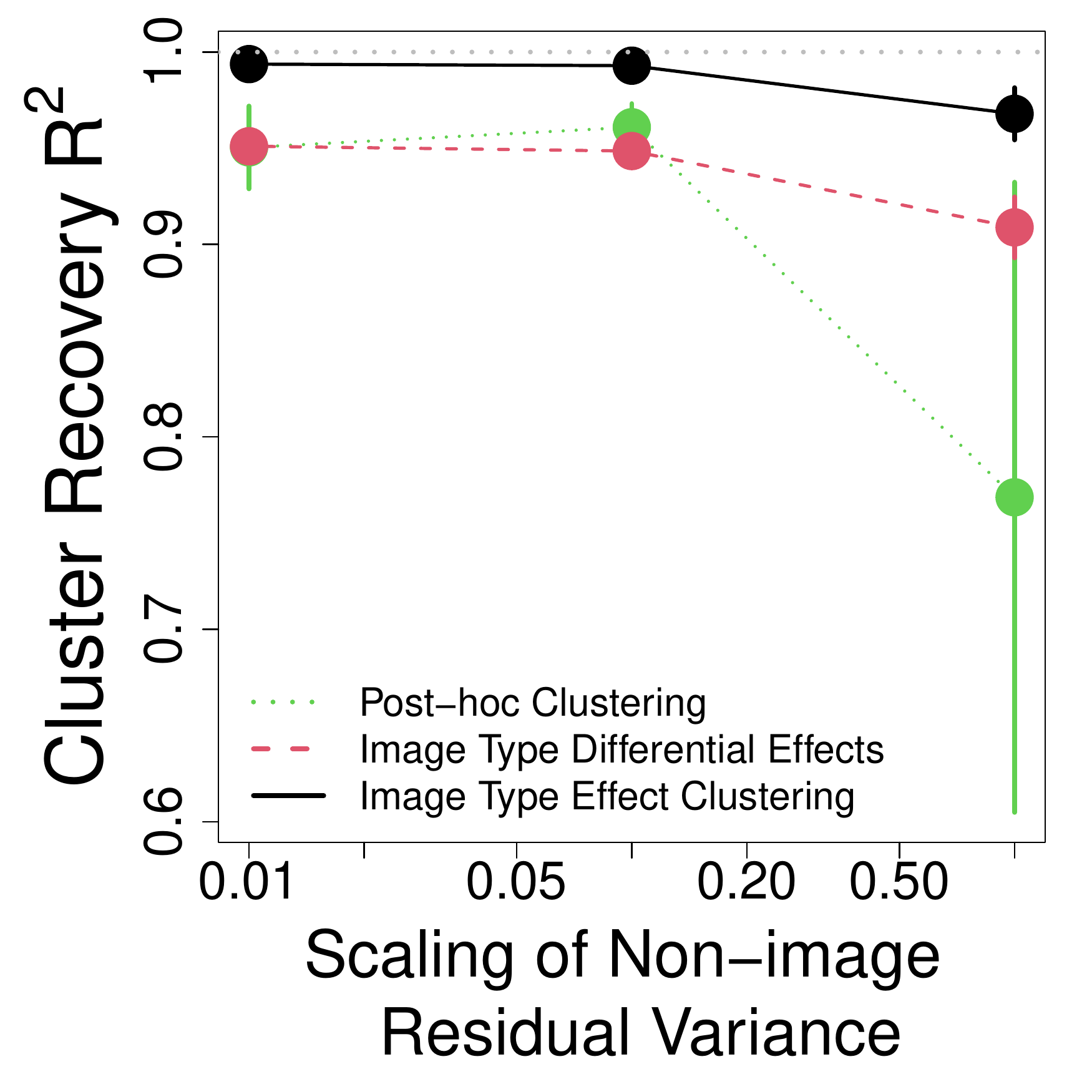}
\end{SCfigure}

\newparagraph{Simulation Results}
We see in the left panel of Figure \ref{fig:SimResults} one representative posterior distribution over $Y_i(1)-Y_i(0)$ given estimated cluster information. We find that the estimated clusters capture the bimodality present in the true distribution of $\tau_i$'s. The right panel shows how the cluster recovery measure for the Image-Type Differential Effects Model and TARNet post-hoc clustering are similar in the low residual variance setting. In the high residual variance setting, the TARNet clustering struggles somewhat in recovering the oracle cluster centers. In contrast, the parsimonious Image-Type Effect Clustering Model performs best at recovering the clustering of the treatment effects across the noise range. This is encouraging for the application of the Image-Type Effect Cluster Model in practice, as presumably, the signal-to-noise ratio for real tasks involving earth observation images is relatively high. 

\section{Application to an Anti-Poverty Experiment in Uganda}\label{ss:app}
\newparagraph{Data} In our application, we explore the effects of the anti-poverty experiment performed in Uganda and described in \S\ref{s:intro}. The treatment variable is the random assignment of small teams to receive grants for business ventures. The outcome variable is an aggregate summary of skilled labor (see \ref{ss:AppOutcome}) measured at the end of the experiment (two years after treatment assignment). De-identified outcome and treatment data were given voluntarily by subjects and are available under CC0 1.0 license. Longitude/latitude information about respondents' villages is found using OpenStreetMap. 

Pre-treatment image data are taken from Landsat. We use the Orthorectified ETM+ pan-sharpened data product, processed to contain minimal cloud cover. Reflectance is measured in the green, near-infrared, and short-wave infrared bands. These bands are useful in capturing information about peak vegetation, water content, and thermal dynamics, in addition to structural land features. 

\begin{figure}[ht]
 \begin{center}  
\includegraphics[width=0.69\linewidth]{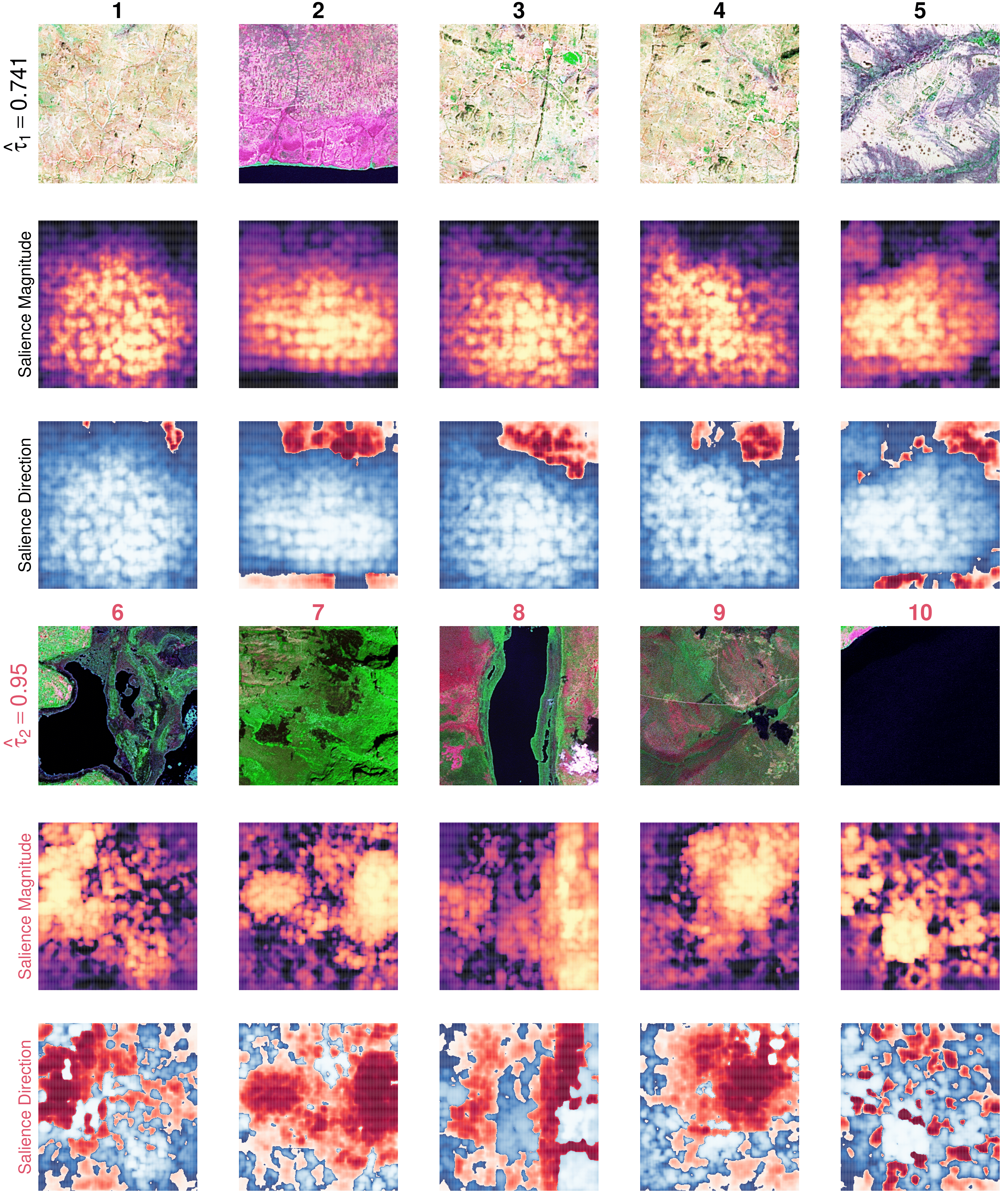}
\includegraphics[width=0.30\linewidth]{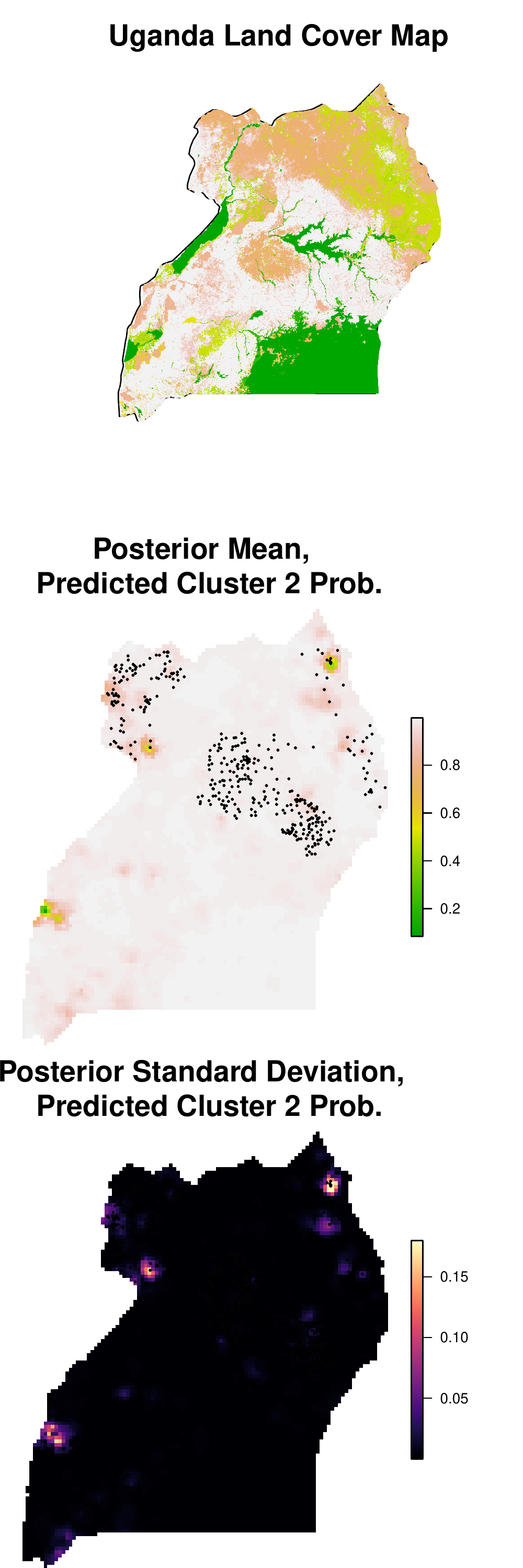}
\caption{\emph{Left, top 3 rows:}  High  probability cluster 1 images. \emph{Left, bottom 3 rows:} High probability cluster 2 images. ``Salience Magnitude'' and ``Direction'' are defined in  \S\ref{ss:sensitivity}. 
\emph{Right, top:} A land use map of Uganda.
\emph{Right, center:} Posterior predictive mean cluster 2 probabilities for the entire country. Black circles represent observed sample points. \emph{Right, bottom}. Posterior standard deviation of the cluster 2 probabilities.
}\label{fig:ImageExemplars}
\end{center}
\end{figure}

\newparagraph{Empirical Results} Due to space constraints, we focus on results from the Image-Type Effect Cluster Model (details in \S\ref{ss:ImplementationDetails}). We set the cluster number to 2 after finding that cluster probabilities become highly correlated with additional clusters. The top three rows in the left panel of Figure \ref{fig:ImageExemplars} show results for the highest images having the highest posterior mean probabilities for cluster 1. For each image, this figure visualizes the salience measures defined in \S\ref{ss:sensitivity}. The bottom portion shows results for the highest posterior mean cluster 2 images. The effect for cluster 1 is  substantially different than for cluster 2. Visually, we see that smaller effects exist for places with harsher terrain and less developed transportation networks, hampering economic growth. These low responders are found in the harsh mountainous northern part of Uganda. 
This is logical, as skilled labor tends to thrive in areas that are connected via transportation networks \citep{ashraf2011cultural}.

We show in the right panel of Figure \ref{fig:ImageExemplars} how the results of the experiment may be generalized to the entire country of Uganda, assuming no systematic bias in places chosen conditional on the image information. In particular, we show the posterior predictive mean cluster 2 probability for the entire country. This kind of analysis can provide policymakers with potentially useful information for how to improve the targeting of treatments in the future across larger geographic contexts.

The Appendix contains supplementary analyses. For example, 
we show in Figure \ref{fig:XMCor} the correlation between the estimated Image CATEs and Tabular CATEs using various conditioning sets, as well as, in Table \ref{tab:CorTab}, between the cluster probabilities and other individual-level covariates. We show in Figure \ref{fig:UncertainImages}  the images having the greatest uncertainty in the cluster probabilities (estimated by the posterior standard deviation). In \S\ref{ss:OrthoEmpirical}, we orthogonalize the potential outcomes using tabular information, and the results remain similar: the correlation between raw and non-orthogonalized cluster probabilities is 0.85. Because our results remain similar after orthogonalization, the  satellite images seem to supply independent and thought-provoking information about effect heterogeneity.

\section{Discussion and Conclusion}\label{s:discussion}
Scientists and policymakers use RCTs to estimate population-wide effects (ATE) and sub-population effects (CATE), using tabular data collected at baseline, often near-time to when the RCT is launched. However, these near-time variables tend to miss important historical or neighborhood-level features. While such features are often unavailable or expensive to collect, satellite images are a data stream that captures such characteristics in an unstructured form. As no CATE method exists explicitly for image analysis, this paper presents principles and modeling strategies for analyzing image-based CATE using probabilistic image-type models. After deriving some model properties, we perform approximate inference using variational methods. Dynamics are explored via simulation; an anti-poverty field experiment from Uganda is analyzed, where we seem to find interesting heterogeneity.

Our approach has limitations, which serve to motivate future research. First, our models estimate heterogeneity clusters at the image level, but not explicitly for smaller segments of an image. Having such within-image heterogeneity segmentation would further improve understanding of what in the image is generating heterogeneity. Second, our methods estimate heterogeneity with respect to a fixed baseline (i.e., the control intervention). While the choice of baseline is clear in most settings, in unclear cases, investigators may need to explore different baselines and compare results. Third, our model is  tailored for RCTs (i.e., assuming unconfoundedness); more research is required to adapt it for observational settings. Using experimental data, effect estimates are confounding-free by design;  heterogeneity can be studied independently of identification. Observational data are more plentiful but require adjustment~\citep{rosenbaum2010design}. We have also viewed images from solely the perspective of surrogate effect modification (in the language of \citet{hernanbook}); the use of images as causal modifiers or mediators is left for future study. 

%

Finally, our focus on \textit{CATE for images} opens exciting possibilities. It not only encourages others to start incorporating satellite images in their planned experiments but also reanalyze past experiments, potentially unraveling previously undetected yet significant sources of effect heterogeneity. As demonstrated in our analysis of the Ugandan anti-poverty experiment, our method identifies heterogeneity not initially detected by incorporating informative satellite data. Thus, our image-based methods have the potential to contribute to policy by complementing traditional RCT heterogeneity analysis based on tabular $\bX_i$---and to analyses in other fields such as agriculture, disaster relief, climate science, and medicine where image data are also prevalent. \hfill $\square$

\section{Acknowledgements} 
The authors thank James Bailie, Cindy Conlin, Devdatt Dubhashi, Felipe Jordan, Mohammad Kakooei, Eagon Meng, Xiao-Li Meng, Markus Pettersson, as well as seminar participants at the Causal Data Science Meeting, Texas Methods Workshop, and RAND CCI Symposium for valuable feedback on this project. We also thank Xiaolong Yang for excellent research assistance. 

\bibliographystyle{plainnat}
\bibliography{heterogeneitybib}


\renewcommand{\thefigure}{A.\arabic{figure}}
\setcounter{figure}{0}  

\renewcommand{\thetable}{A.\arabic{table}}
\setcounter{table}{0}  

\renewcommand{\thesection}{A.\arabic{section}}
\setcounter{section}{1}


\section*{Appendix}

\subsection{Open-source Software \& Reproducability}
We make the modeling strategies introduced in this paper accessible in an open-source software package available at \Verb|github.com/cjerzak/causalimages-software|. For an up-to-date tutorial regarding package use, see \Verb|github.com/cjerzak/causalimages-software#readme|. Replication data for the experiment analyzed in the application are contained in this \Verb|GitHub| repository as well (we include both the experimental data from the original investigators and the geo-referenced satellite images).

\subsection{Supplementary Information for the Image-Type Probabilistic Models}\label{ss:ModelDetails}
 

\subsubsection{Deriving the Conditional Distribution, $\{\tau_i = Y_i(1)-Y_i(0)|Z_i=z\}$}\label{ss:DistributionTau}
Using the model outlined in the main text, conditioning on $\tau_i$, and exploiting Normality,
\begin{align*}
\{ Y_i(1) - Y_i(0) \mid Z_i = z, \mu_{\tau_i} \} \sim \mathcal{N}(\mu_{\tau_i}, \ \sigma_{0,z}^2 + \sigma_{1,z}^2)
\end{align*}
Integrating out $\mu_{\tau_i}$:
\begin{align*}
p(Y_i(1) - Y_i(0) = \tau_i \mid Z_i = z) &= \int_{-\infty}^{\infty} \ p(Y_i(1) - Y_i(0) = \tau_i \mid Z_i = z, \ \mu_{\tau_i})  \ p(\mu_{\tau_i} \mid Z_i = z) \; \textrm{d}\mu_{\tau_i}\\ &= \int_{-\infty}^{\infty}
\frac{1}{\sqrt{2\pi (\sigma_{0,z}^2 + \sigma_{1,z}^2)}} \exp\bigg\{
\frac{-(\tau_i-\mu_{\tau_i})^2}{2 (\sigma_{0,z}^2 + \sigma_{1,z}^2) }
\bigg\} 
\\ &\qquad \qquad \qquad \times \frac{1}{\sqrt{2\pi \sigma_{\tau,z}^2}} \exp\bigg\{
\frac{-(\mu_{\tau_i}-\mu_{\tau,z})^2}{2\sigma_{\tau,z}^2 }
\bigg\} \; \textrm{d}\mu_{\tau_i}
\\ &= \frac{1}{\sqrt{2 \pi ([\sigma_{0,z}^2 + \sigma_{1,z}^2] + \sigma_{\tau,z}^2)}} \exp \bigg\{ 
\frac{-(\tau_i - \mu_{\tau,z})^2}{2 ([\sigma_{0,z}^2 + \sigma_{1,z}^2] + \sigma_{\tau,z}^2)}
\bigg\}. 
\end{align*}
Therefore, 
\begin{align*}
\{\tau_i = Y_i(1) -Y_i(0) \mid Z_i=z\} \ \sim \ \mathcal{N}\left(\mu_{\tau,z}, \ \sigma_{0,z}^2+\sigma_{1,z}^2+\sigma_{\tau,z}^2\right).
\end{align*}

\subsection{Simulation Details}
\begin{figure}[ht]
 \begin{center}  
\includegraphics[width=0.95\linewidth]{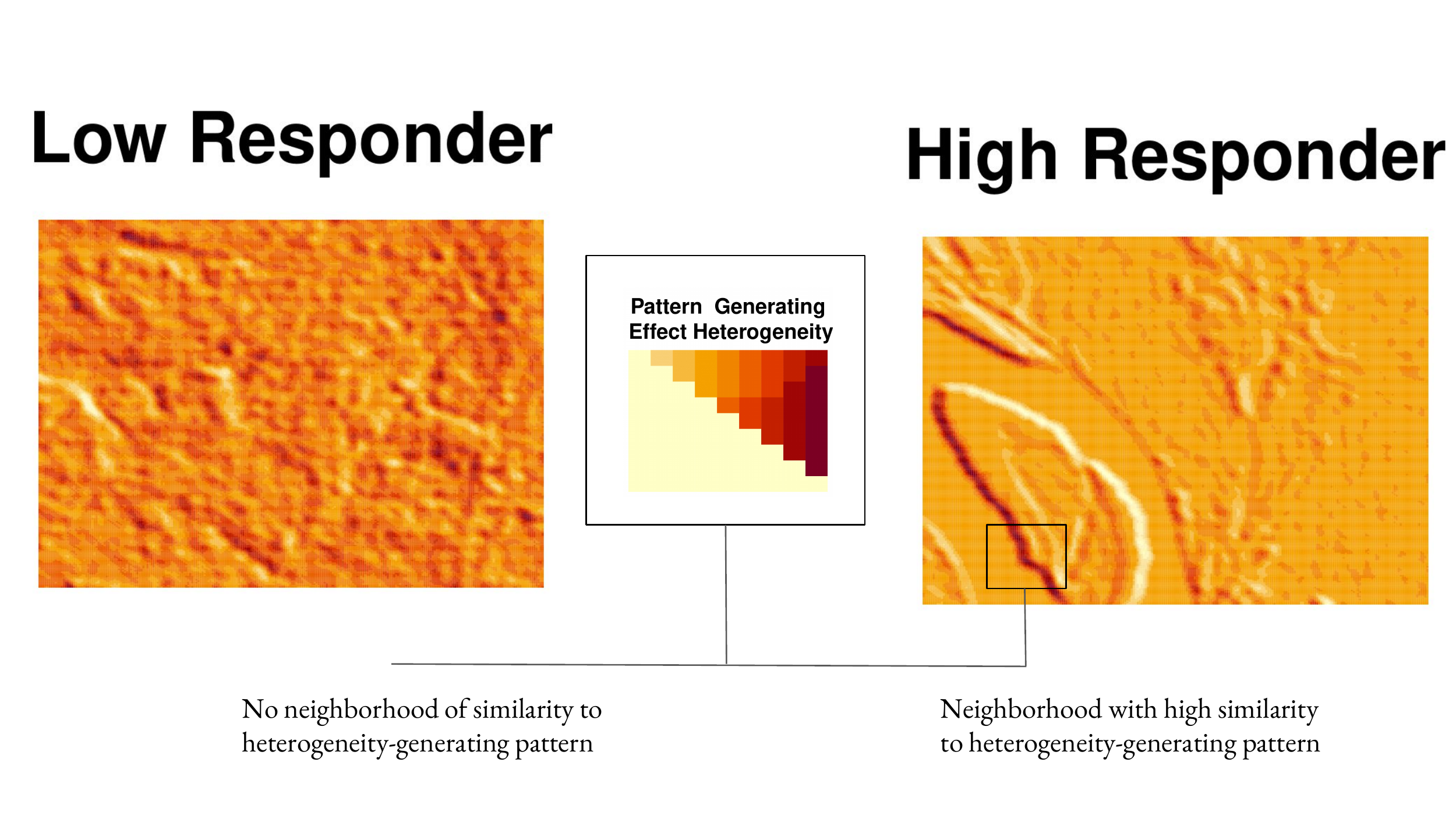}
\caption{Simulation design illustration. \emph{Center:} The image pattern used in generating the heterogeneity response in the simulation design of \S\ref{ss:sim}. \emph{Left:} An image having no regions of strong similarity to the heterogeneity-generating pattern (leading to a low treatment effect). \emph{Right:} An image with many regions of strong similarity to the heterogeneity-generating pattern (leading to a high treatment effect).
}
\label{fig:SimIll}
\end{center}
\end{figure}

\begin{SCfigure}
\includegraphics[width=0.45\linewidth]{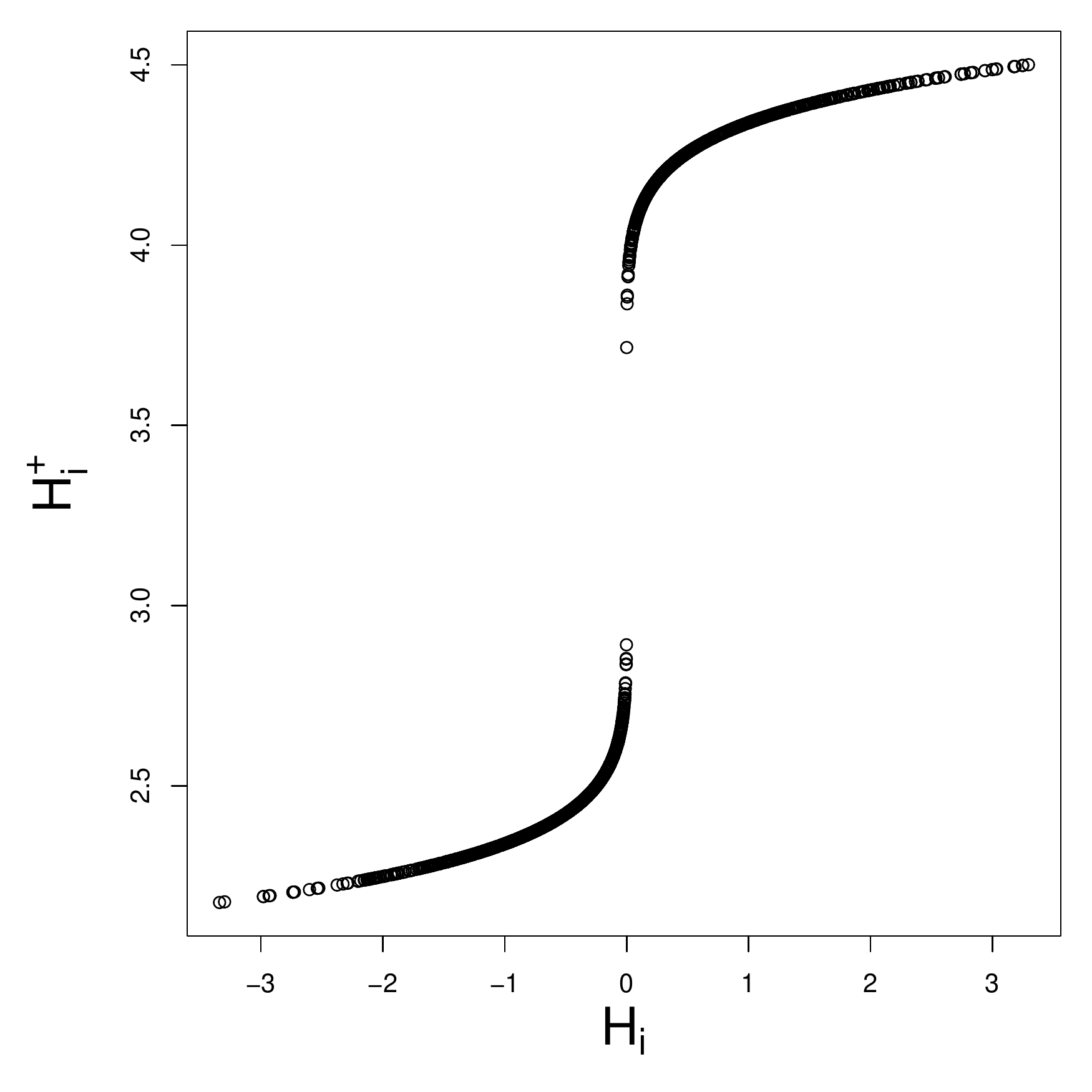}
\caption{Illustration of the non-linear transformation used in the simulation for generating $H_i^{+}$ from $H_i$.}
\end{SCfigure}

\clearpage 

\subsection{Supplementary Analyses for the Application}\label{ss:SuppEmpirical}
\subsubsection{Additional Data Description}\label{ss:AppOutcome}
We obtain satellite data for the neighborhood around each experimental unit in the following way. First, the place name of residence associated with each unit was geo-referenced using OpenStreetMap and, of this geo-referencing failed, the Google Geocoding API. When this second geo-referencing attempt failed, we use the geometric center for the layer associated with the geographic unit as our focal point for the given unit. Satellite information was then obtained for a cube around focal points with side lengths of 5000 meters. For the skilled work outcome, we take the scaled sum of the log hours worked by experimental units in the last 7 days in skilled or highly skilled trades. 

Despite our best efforts, there is still room for error in this geo-coding process. We expect that such errors would introduce random noise into the analysis, drowning out potential signal and introducing attenuation bias of conditional effects towards 0. 

\subsubsection{Additional Application Analyses}\label{ss:SuppCorTab}
\begin{figure}[ht]
 \begin{center}  
\includegraphics[width=0.99\linewidth]{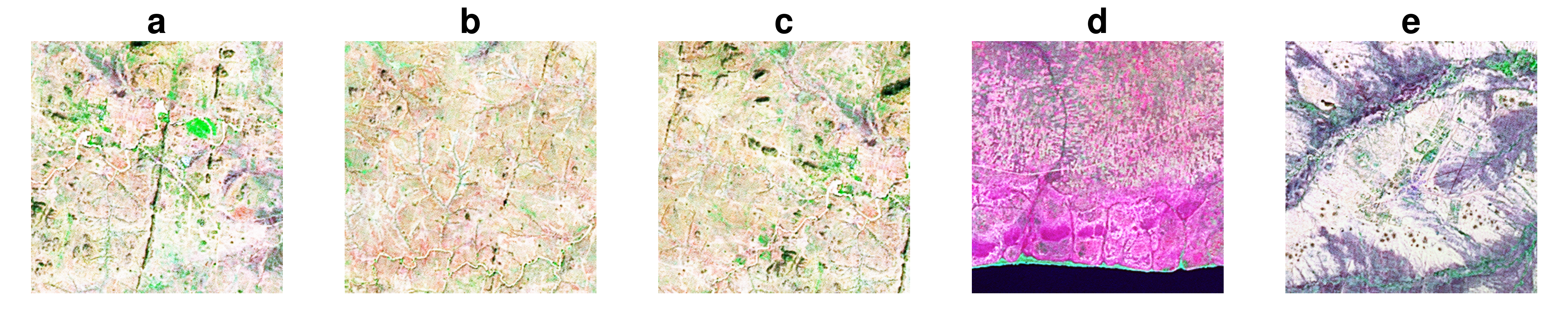}
\caption{ Images with greatest uncertainty in cluster probabilities from the main empirical analysis.}\label{fig:UncertainImages}
\end{center}
\end{figure}

\begin{figure}[ht]
 \begin{center}  
\includegraphics[width=0.9\linewidth]{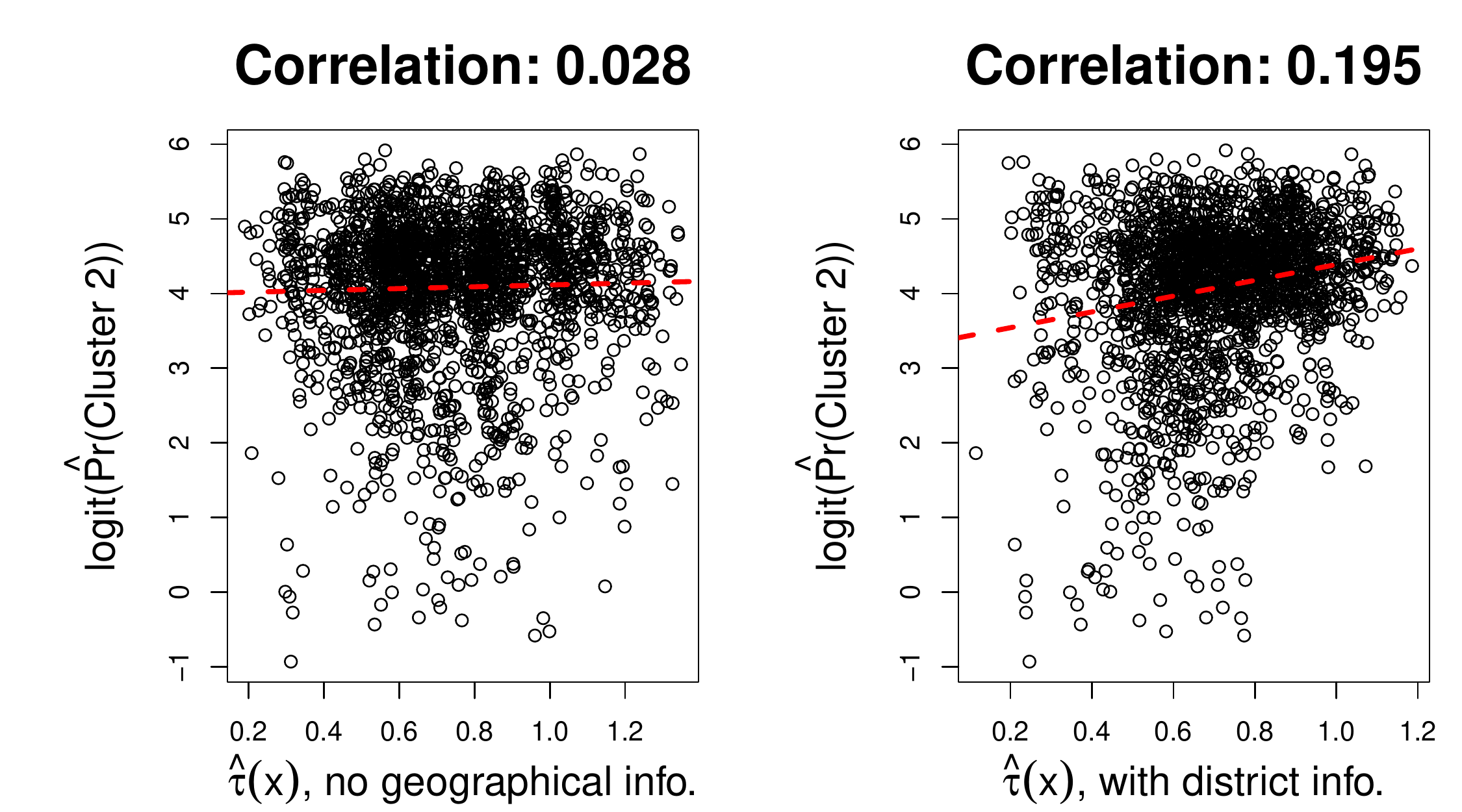}
\caption{\emph{Left}. Correlation between estimated treatment effects using a causal forest with individual-level tabular covariates and the posterior mean cluster 2 probabilities from the image heterogeneity model. Individual-level covariates include gender, education, parental education, and indicators for whether a unit's mother or father were alive at the start of the experiment. \emph{Right}. Correlation between estimated treatment effects using a causal forest with individual-level tabular covariates along with district-level indicators and the posterior mean cluster 2 probabilities. As expected, the correlation increases, but there is still considerable information present in the estimated clusters not reducible to district indicators alone.}\label{fig:XMCor}
\end{center}
\end{figure}

Here, we include additional analyses associated with the main application. In particular,  in Figure \ref{fig:XMCor}, we see the correlation between the estimated Image CATEs and Tabular CATEs using various conditioning sets. These correlations are further broken down by tabular covariate in Table \ref{tab:CorTab}. We show in Figure \ref{fig:UncertainImages}  the images having the greatest uncertainty in the cluster probabilities (estimated by the posterior standard deviation).

\input{./CorTabFALSE.tex}

\subsection{Empirical Analysis with Orthogonalized Potential Outcomes}\label{ss:OrthoEmpirical}
To understand the degree to which the neighborhood-level satellite information is a proxy for tabular covariate information, we perform an image CATE analysis in the space of orthogonalized potential outcomes. We orthogonalize potential outcomes by fitting a model for the observed potential outcomes using tabular covariates and residualizing. The potential outcome model here used is a linear regression model predicting each observed outcome using main treatment effects and interactions between treatment and gender, treatment and baseline human capital, and treatment and baseline business capital (as well as the main effect terms for the associated interaction).  We use this functional form because it is similar to a model used in the original experimental analysis. We find an absolute correlation of 0.85 between the cluster probabilities using the orthogonalized and raw outcomes.

\subsection{Model Implementation Details}\label{ss:ImplementationDetails}
In the implementation of our models using Bayesian CNN arms, we leave the number of hidden layers, the filter size, the stride length and other quantities as hyper-parameters that can be set by investigators. Future work should explore the implications of these choices on the practical considerations of probabilistic causal image analysis. That said, there are some general principles that are evident from prior research, such as the idea that, with more data, the number of hidden layers can be increased.

We also here add information about the choice of priors in the Bayesian model. The unconstrained components of the uncertainties are drawn from Gaussians with mean and variance scaled indexed to $z$; the non-negativity of the variance is enforced through the softplus transformation (where softplus$(x)$ = $\log(1+\exp(x))$). Neural network parameters receive priors using the Empirical Bayes' approach described in \citet{krishnan2020specifying}. We have found the use of Empirical Bayes to be important practice: given the highly non-linear parametric functions applied here, seemingly non-informative priors in the parameter space (e.g., $\mathcal{N}(0,10)$) can be highly informative in the space of induced transformations. The prior for the treatment effect mixture components are centered around the non-parametric difference-in-means estimator for the ATE. 

In our application, we use four convolutional layers (filter dimension 5$\times$5), separated by max-pooling layers (2$\times$2). Each convolutional layer applies 32 filters. Bottleneck projection layers are used after each convolutional layer, projecting the 32 dimensions down to 3 to keep the number of parameters reasonably low given the small sample size available in the application. Batch normalization layers are used across the feature dimension after each non-linearity (batch normalization momentum across each update step is $=$ 0.90). The swish activation is used. We apply the Gumbel-Softmax to approximate the random categorical sampling with the temperature parameter set to 0.5. We use the flipout estimator for the neural parameter sampling \citep{wen2018flipout}. Five Monte Carlo iterations are used in each variational inference training step. With this model structure, each batch sample of 20 units takes about one second on a single Apple M1 GPU using Metal-optimized TensorFlow 2.11. The full simulation suite takes about 12 hours on local hardware.  

\end{document}

%% file: CorTabFALSE.tex
\begin{table}[!htbp] \centering 
  \caption{Correlation of estimated image cluster 1 probabilities with key tabular covariates.} 
  \label{tab:CorTab} 
\footnotesize 
\begin{tabular}{@{\extracolsep{5pt}} cc} 
\\[-1.8ex]\hline 
\hline \\[-1.8ex] 
 & Correlation \\ 
\hline \\[-1.8ex] 
Urban & -0.27 \\ 
Longitude & 0.31 \\ 
Latitude & -0.40 \\ 
Female indicator & 0.01 \\ 
Human capital score & -0.01 \\ 
\hline \\[-1.8ex] 
\end{tabular} 
\end{table}